\documentclass{article}
\usepackage{iclr_conference,times}

\usepackage{amsmath,amsfonts,bm}

\def\eqref#1{equation~\ref{#1}}

\def\1{\bm{1}}

\DeclareMathAlphabet{\mathsfit}{\encodingdefault}{\sfdefault}{m}{sl}
\SetMathAlphabet{\mathsfit}{bold}{\encodingdefault}{\sfdefault}{bx}{n}

\usepackage{hyperref}
\usepackage{url}
\usepackage{graphicx}
\usepackage{booktabs}
\usepackage[table]{xcolor}
\usepackage{arydshln}
\usepackage{wrapfig}

\title{FACT: Fidelity-Aware Construction of Articulated Twins}
\author{
Kuixiang Shao \\
ShanghaiTech University \\
Shanghai, China \\
\texttt{shaokx2025@shanghaitech.edu.cn}
\And
Chuansen Nie \\
ShanghaiTech University \\
Shanghai, China \\
\texttt{niechs2025@shanghaitech.edu.cn}
\And
Yinuo Bai \\
ShanghaiTech University \\
Shanghai, China \\
\texttt{baiyn2026@shanghaitech.edu.cn}
\And
Jiayuan Gu \\
ShanghaiTech University \\
Shanghai, China \\
\texttt{gujy1@shanghaitech.edu.cn}
\And
Jingyi Yu \\
ShanghaiTech University \\
Shanghai, China \\
\texttt{yujingyi@shanghaitech.edu.cn}
}

\iclrfinalcopy
\begin{document}
\maketitle

\begin{abstract}
Visually plausible articulated assets may still fail during contact interactions or exhibit inaccurate motion. We present \textbf{FACT} (Fidelity-Aware Construction of Articulated Twins), an agentic framework that progressively constructs articulated twins to improve geometry, contact, and dynamic fidelity. The agent drives an evidence--diagnosis--revision loop on a shared editable representation, selecting measurements and model edits using quantitative feedback, while numerical tools execute and validate the updates. It reconstructs editable articulated geometry from images through feature planning, targeted measurements, and diagnostic refinement. On this reference, it repairs collision proxies through task-aware local repartitioning before fidelity-constrained compression. Finally, it constructs response models from passive-response videos, using simulation residuals to guide model revision and constrained physical parameter fitting. Experiments show that FACT improves geometric reconstruction over baselines, enables more reliable interaction with simpler collision proxies, and better reproduces held-out physical responses than direct parameter inference. Project page: \url{https://l-avenir.github.io/fact-project-page/}.
\end{abstract}

\begin{figure}[!ht]
    \centering
    \vspace{-4mm}
    \includegraphics[width=\linewidth]{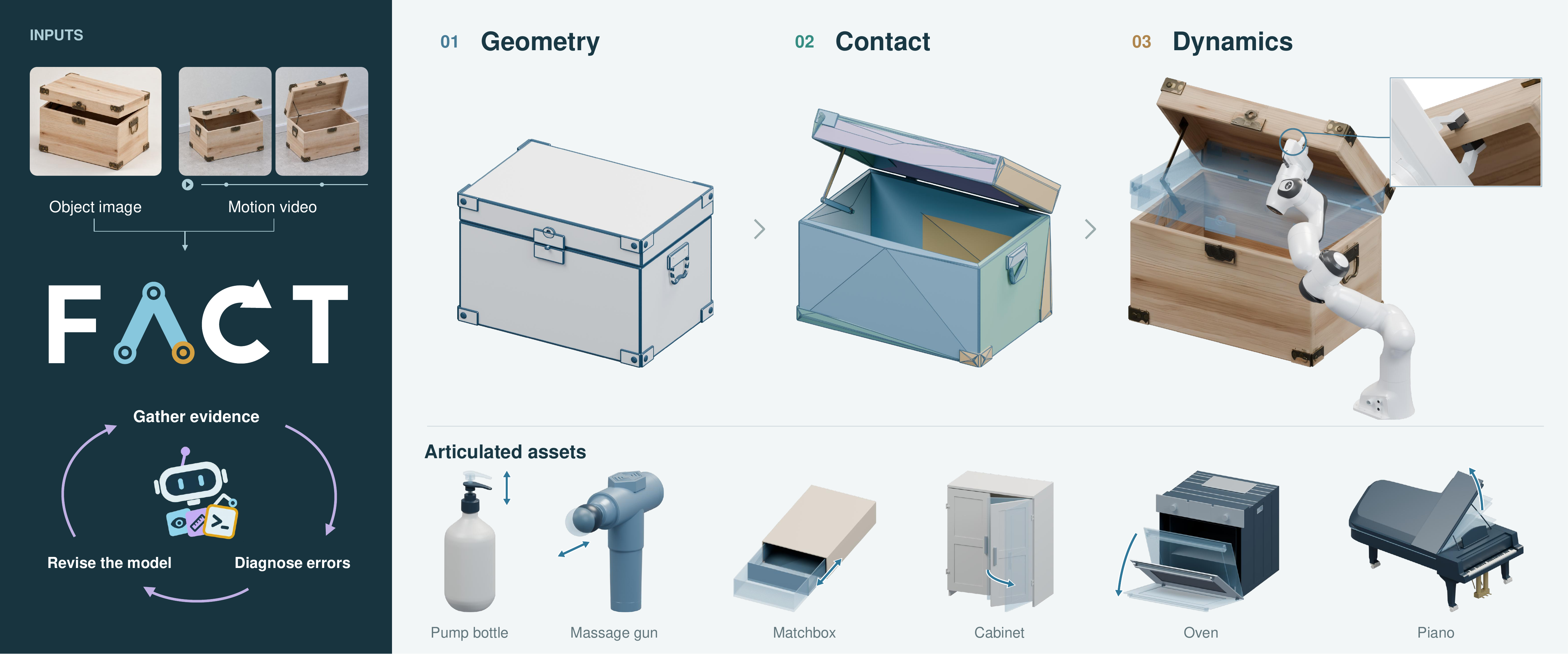}
    \vspace{-6mm}
    \caption{\textbf{Geometry, contact, and dynamics.}  \textbf{FACT} reconstructs editable articulated objects, refines collisions to preserve free space for interaction, and calibrates motion dynamics from video.}
    \vspace{-4mm}
    \label{fig:teaser}
\end{figure}

\section{Introduction}
\label{sec:intro}

Simulation enables embodied agents to gain experience through repeated interactions with diverse objects. Advances in simulation platforms, articulated asset collections, and procedural generation have broadened this opportunity~\citep{Xiang_2020_SAPIEN,geng2023gapartnet,joshi2025articulated}, while generative and agentic systems make asset creation from visual inputs increasingly accessible~\citep{le2025articulate,zhou2026articraft,wang2026embodiedgen}. Yet increasing the number of visually plausible articulated assets does not by itself provide more faithful interaction experience. An asset determines both what a robot observes and what happens when it acts: where contact can be established, which motions are feasible, and how the object's motion evolves over time. Building useful simulation-ready assets therefore requires capturing interaction feasibility and motion responses alongside shape and articulation.

This gap is particularly clear in automatic articulated asset creation. Articraft generates and refines programs for parts and joints from images~\citep{zhou2026articraft}. However, image-based reconstruction can still misrepresent drawer depth, handle clearance or interior cavities, yielding an asset that looks plausible but does not support the intended interaction. Even when these structures are accurate, collision approximation can compromise otherwise feasible interactions. Pipelines such as EmbodiedGen V2 use CoACD to obtain convex collision proxies~\citep{wang2026embodiedgen,wei2022coacd}. Although CoACD accounts for collision-related geometric error, proxies can still fill handle openings or obstruct drawer travel. Yet contact feasibility alone does not ensure a faithful motion response: Articraft uses language-model priors for mass and damping, whereas reproducing instance-specific responses calls for calibration against observed trajectories. These considerations motivate three complementary aspects of fidelity: \emph{geometry fidelity} in structure and dimensions, \emph{contact fidelity} in task-specific interaction and motion feasibility, and \emph{dynamic fidelity} in motion responses.

Addressing these errors calls for progressively grounding a shared articulated model. Reconstructed geometry and articulation guide collision refinement and, together with the refined proxies, form a fixed basis for dynamic calibration. The appropriate measurement and correction depend on the part and the failure: a narrow opening requires different evidence from an incorrect joint response. An agent drives this evidence--diagnosis--revision loop, selecting measurements and model edits using semantic understanding and quantitative feedback, while numerical tools execute and validate the updates. The central design is to link geometric discrepancies to editable features, free-space and motion violations to collision partitions, and trajectory residuals to response models and physical parameters.

We present \textbf{FACT}, an agentic framework that progressively constructs and calibrates an articulated digital twin (Fig.~\ref{fig:teaser}). First, the agent plans parametric features and selects measurements to constrain their dimensions and placement, using local diagnostics to reconstruct editable geometry and articulation from an image-generated mesh. Keeping this geometry and articulation fixed, it identifies likely interactions and revises local collision partitions to repair free-space and motion violations, then compresses the proxies under fidelity constraints. Finally, with the resulting geometry, joints, and proxies fixed, the agent uses videos to estimate trajectories in the recovered joint coordinates and construct response models. It then uses simulation residuals to revise the models and guide constrained parameter fitting.

Our contributions are threefold:
\begin{itemize}
    \item We introduce an agentic reconstruction method that combines feature planning, targeted measurements, and diagnostic refinement to recover editable articulated geometry.
    \item We develop agent-guided collision refinement that diagnoses task-specific contact failures, revises local partitions, and compresses proxies under fidelity constraints.
    \item We formulate residual-guided dynamic calibration in which the agent autonomously constructs and revises response models and directs constrained system identification from interaction observations.
\end{itemize}

\section{Related Work}
\label{sec:related_work}

\paragraph{Articulated object reconstruction and generation.}
Articulated reconstruction recovers part geometry and kinematics from observations across object states or continuous motion~\citep{jiang2022ditto,jiayi2023paris,weng2024neural,liu2025building,ai2026articulation}, while geometry-based articulation infers part structure and joints from meshes or point clouds~\citep{qiu2025articulate,wang2026artllm,li2026urdf}. Generative approaches learn joint representations of geometry, connectivity, and motion~\citep{lei2023nap,Liu_2024_CAGE,gao2025meshart,su2025artformer}, with image-conditioned methods producing articulated assets from visual cues~\citep{urdformer_2024,liu2025singapo}. Code-based and agentic pipelines express articulated assets as executable models~\citep{zhao2025real2code,le2025articulate,zhou2026articraft}. Complementary work addresses hidden geometry completion~\citep{boudjoghra2026unfoldart}, non-penetration and mobility constraints~\citep{kreber2025guiding}, and physical attributes for simulation~\citep{cao2026physx,yang2026physforge}.

\paragraph{Programmatic mesh reconstruction and agentic 3D modeling.}
Structured geometric representations range from compact assemblies of primitives and superquadrics~\citep{ye2025primitiveanything,fedele2025superdec,tavernini2026superflex} to shape programs built from parameterized parts, assembly relations, and reusable operations~\citep{jones2020shapeAssembly,jones2021shapemod,jones2023ShapeCoder,jones2026shapelib}. Shape-to-code approaches reconstruct executable programs from point clouds or images~\citep{dai2026meshcoder,rukhovich2025cad,kolodiazhnyi2025cadrille}. Program generation is complemented by agentic frameworks that combine structural planning, code execution, and validation~\citep{zhang2026shapecraft,zhou2026articraft,lin2026procedura}, and by visual or geometric feedback for iterative refinement~\citep{Kabisov_2026_CVPR,hu2026itercad}. In \textbf{FACT}, feature plans guide measurements of dimensions and placement, while local diagnostics determine whether to revise a feature representation or its parameters.

\paragraph{Collision geometry for interactive simulation.}
Collision proxies can be constructed through approximate convex decomposition of non-convex meshes~\citep{mamou2016volumetric,wei2022coacd}. Decomposition strategies include learned cutting policies, visibility-based partitioning, and neural feature fields~\citep{RL-ACD,fokin2026visacd,learningconvexdecomp2026}. Application requirements also guide decomposition: methods preserve navigable space~\citep{andrews2024navigation} or maintain coherent decompositions across animation poses~\citep{Thul2018AACD}. \textbf{FACT} uses CoACD as a backend, with task and motion diagnostics guiding local repartitioning and subsequent compression under fidelity constraints.

\paragraph{Physical digital twins and system identification.}
Visual system identification couples differentiable simulation and rendering to infer physical parameters from videos~\citep{gradsim,li2023pacnerf}, while deformable digital twins recover geometry and material response from observed interactions~\citep{zhong2024springgaus,jiang2025phystwin}. Robot-interaction pipelines instead use torque sensing, proprioception, and motion observations to estimate object properties~\citep{pfaff2025scalable,chen2025learning,lou2026d}. RigPI combines visual priors with torques and poses to identify inertial and frictional parameters of articulated objects~\citep{he2026rigpi}. In \textbf{FACT}, the agent autonomously constructs and revises response models from observed motion and directs constrained numerical fitting.

\section{Method}
\label{sec:method}

\subsection{Overview}
\label{sec:method_overview}

\begin{figure}[ht]
    \centering
    \vspace{-2mm}
    \includegraphics[width=\linewidth]{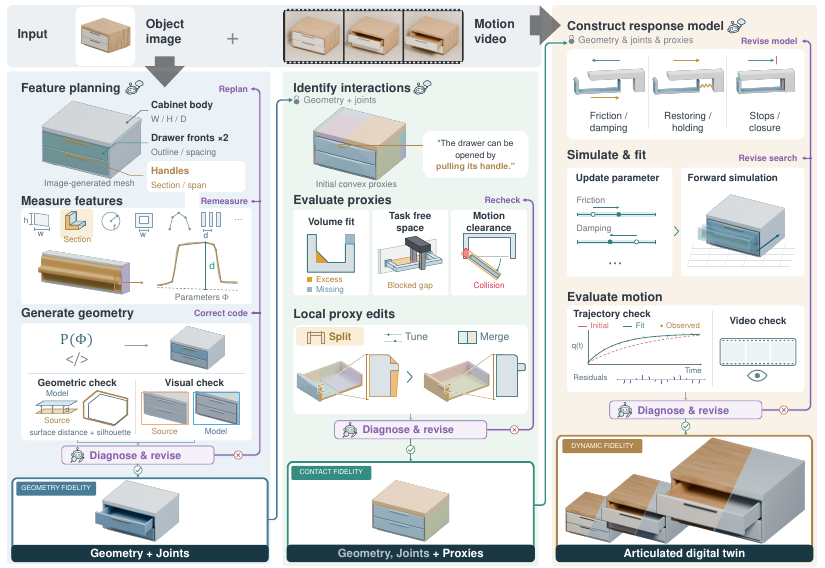}
    \vspace{-6mm}
    \caption{\textbf{FACT overview.} The agent drives an evidence–diagnosis–revision loop on a shared articulated representation. It plans geometric features and selects measurements, directs task-aware collision repair followed by fidelity-constrained compression, and constructs and revises response models while guiding constrained parameter fitting.}
    \vspace{-6mm}
    \label{fig:method_overview}
\end{figure}

\textbf{FACT} constructs an articulated twin from object images and motion videos (Fig.~\ref{fig:method_overview}). We represent the twin as $\mathcal{A}=(G,K,C,\theta)$: part geometry, a kinematic graph, collision proxies, and physical parameters. Feature planning, measurement, refinement, and articulation establish $(G,K)$ (Sec.~\ref{sec:method_geometry}). Task-aware contact refinement repairs $C$ on fixed $(G,K)$ before compression under volumetric, free-space, and sampled-motion constraints (Sec.~\ref{sec:method_contact}). Dynamic calibration constructs and revises a passive response model and fits $\theta$ to video trajectories on fixed $(G,K,C)$ (Sec.~\ref{sec:method_dynamic}). Each stage follows an evidence--diagnosis--revision loop: the agent selects measurements and model revisions, while numerical tools execute these operations and return quantitative feedback. Shared part identities and body frames maintain correspondence across geometry, collision proxies, and dynamic.

\subsection{Geometry Fidelity: Feature Planning, Measurement, and Refinement}
\label{sec:method_geometry}

We separate feature planning from numerical measurement, using reconstruction feedback to revise the feature representation or its parameters.

\paragraph{Feature planning.}
The agent inspects whole-object and isolated part views of an image-generated mesh $M$ to identify logical parts and their structural requirements. A feature plan specifies feature types, relations, parameters to measure, and links to agent-identified source regions. Each part may combine extrusions, sweeps, shells, and subtractive features, with custom constructions for shapes outside this vocabulary. Features are associated with the visible structures they must preserve, including openings, contacts, and repeated elements. Planning determines what to represent and measure, leaving numerical values and measurement procedures to the next step.

\paragraph{Feature-conditioned measurement.}
For each planned parameter, the agent selects a source region, coordinate frame, and measurement procedure; geometric tools compute the values. Projected contours and axial extents constrain an extrusion's profile and depth, whereas cross-section centers and contact regions constrain a handle's path and endpoints. Measured contours are reduced to editable profile controls. Dependent parameters are derived from measurements using symmetry, repetition, and contact relations. We retain each parameter's measurement or derivation for later revision. The agent implements the plan as an executable program $P(\phi)$ with geometric parameters $\phi$, which generates reconstructed part meshes $\widehat{G}$ from editable features rather than copied source triangles.

\paragraph{Diagnostic refinement.}
We compare $\widehat{G}$ with $M$ using bidirectional surface distances, matched-view silhouettes, and side-by-side renders. Part-level diagnostics use agent-identified source correspondences to localize errors, while targeted views or sections help resolve structures obscured in standard views. The agent traces discrepancies to the responsible modeling decision: an inadequate feature representation triggers replanning, incorrect dimensions or placement trigger remeasurement, and implementation errors trigger program correction. The revised program regenerates the meshes for reevaluation. A reconstruction is accepted only when it passes both automated geometric validation and the agent's visual inspection.

\paragraph{Functional completion and articulation.}
After validation, a separate pass adds or refines minimal functional structures, such as cavities, supports, and mounts. Category knowledge guides unobserved structures, while measured envelopes and functional clearances constrain their dimensions. Parts that move together are grouped into rigid bodies, and each joint in $K$ specifies its parent and child bodies, type, axis, origin, and motion limits. Sampled poses, including travel endpoints, reveal interferences that prompt geometric or joint-placement corrections. The resulting $G$ and $K$ form the articulated reference for contact refinement.

\subsection{Contact Fidelity: Task-Aware Collision Refinement}
\label{sec:method_contact}

We repair task-dependent collision errors before compressing the proxy under fixed quality constraints.

\paragraph{Collision representation and initialization.}
For each rigid body $b$, let $G_b$ denote the region occupied by its reference geometry. Its collision proxy is $C_b=\bigcup_{h=1}^{H_b}C_{b,h}$, a union of $H_b$ convex hulls. We use $C$ to denote the full collection of hulls across bodies. Decomposition regions serve as inputs to CoACD~\citep{wei2022coacd} and may be repartitioned within a body while retaining all source geometry. Reference geometry $G$ and articulation $K$  remain fixed. A complete default regional decomposition $C^0$ initializes the proxy and sets the total hull budget $H_{\max}=H(C^0)$.

\paragraph{Task-aware diagnostics.}
The agent identifies likely interactions and selects contact regions, access spaces, and motion clearances for evaluation. Task free space is defined outside the union of all reference body geometries at each pose, independently of the candidate proxy. We evaluate per-body volumetric overlap, missing and excess volume, task-region occupancy, and false interbody collisions at reference-confirmed collision-free poses. Motion checks sample the full range, including endpoints, with denser sampling near failures; diagnostics identify the responsible hull pairs and source regions.

\paragraph{Diagnostic partition refinement.}
Based on these diagnostics, the agent selects a local edit: repartitioning or splitting a region, adjusting regional decomposition parameters, or merging compatible regions or hulls within a body. Separating a handle into side supports and a bridge, for example, can prevent convex hulls from filling its opening. Geometric tools execute the edit and regenerate affected proxies, followed by whole-object reevaluation. Refinement seeks a proxy within the hull budget that passes the sampled motion checks while bounding increases in volumetric and free-space errors relative to $C^0$. Once found, this proxy is fixed as the quality reference $C^a$.

\paragraph{Feasibility-first compression.}
Starting from $C^a$, the agent searches through local edits for a simpler proxy, prioritizing total hull count $H$ over total vertex count $V$:
\begin{equation}
\begin{aligned}
\operatorname*{lexmin}_{C}\quad & \bigl(H(C),V(C)\bigr) \\
\text{subject to}\quad
& C\in\mathcal{F}(G,K),\quad H(C)\leq H_{\max}, \\
& \mathbf{e}(C)\preceq\mathbf{e}(C^{a})+\boldsymbol{\delta}.
\end{aligned}
\label{eq:contact_compression}
\end{equation}
Here, $\mathcal{F}(G,K)$ contains valid proxies with no self-collisions over the articulated motion space. The error vector $\mathbf{e}$ collects per-body $1-\mathrm{IoU}$, missing and excess volume ratios, and free-space occupancy in each selected region. The tolerance vector $\boldsymbol{\delta}$ bounds the componentwise error increases relative to $C^a$. An edit is accepted only if it improves the lexicographic objective and satisfies these constraints.

\subsection{Dynamic Fidelity: Video-Based Calibration}
\label{sec:method_dynamic}

Given one or more videos, we calibrate an object's dynamics with the reconstructed articulated twin $(G,K,C)$ fixed.

\paragraph{Motion-grounded parameterization.}
The recovered kinematic model defines the joint types and coordinate system used to represent the observed motion. Within this representation, the agent selects motion-estimation procedures and invokes analysis tools to estimate time-aligned joint trajectories and initial states. It uses the recovered mechanism, visual cues, and observed motion to autonomously construct the response model, specifying its functional form and initializing its physical parameters $\theta$. Depending on the mechanism, $\theta$ may include parameters governing damping, friction, and state-dependent restoring or holding terms. The parameterization preserves the physical relationships among these quantities and the reconstructed geometry.

\paragraph{Simulation-based inference.}
For observed trajectories $\{q_k^{\mathrm{obs}}\}_{k=1}^{m}$, $m\geq1$, forward simulation predicts $q_k^{\mathrm{sim}}(\theta;\xi_k)$, where $\xi_k$ denotes sequence-specific initial conditions. The agent defines a constrained search space $\Omega$, selecting which physical parameters and uncertain initial-state components to optimize, together with their admissible bounds. Within each fitting step, the response-model form remains fixed, and unselected quantities retain their current estimates.
\begin{equation}
\min_{(\theta,\{\xi_k\})\in\Omega}
\sum_{k=1}^{m}
\left[
\mathcal L_q\!\left(
q_k^{\mathrm{sim}}(\theta;\xi_k),
q_k^{\mathrm{obs}}
\right)
+\mathcal R_k(\xi_k)
\right].
\label{eq:dynamic_calibration}
\end{equation}
Here, $\mathcal L_q$ measures time-aligned trajectory discrepancies, while $\mathcal R_k$ constrains any optimized initial-state components around their observed estimates; it is omitted when initial conditions are fixed. Physical parameters are shared across sequences of the same joint and configuration. The numerical solver returns fitted values, simulated trajectories, and residuals.

\paragraph{Residual-guided refinement.}
The agent uses residual patterns to revise the response-model form and the optimization variables and bounds in $\Omega$. Speed-dependent discrepancies prompt examination of damping, while incorrect stopping or holding behavior prompts examination of resistance terms. When discrepancies suggest uncertain observations or initialization, the agent rechecks the source video or revises which initial-state components are fitted. Numerical tools solve the revised problem, and forward simulation evaluates the resulting motion. After calibration, the response model and physical parameters are fixed for evaluation on held-out clips.

\begin{table}[ht]
    \vspace{-4mm}
    \centering
    \caption{\textbf{Quantitative comparison of geometry fidelity.} Gray rows show upstream Rodin meshes for reference; bold marks the best reconstruction results within each setting.}
    \label{tab:geometry_fidelity_main}
    \small
    \setlength{\tabcolsep}{3pt}
    \begin{tabular}{lcccccc}
        \toprule
        Method & CD$\downarrow$ & SD-P95$\downarrow$ & F@1\%$\uparrow$ & F@2\%$\uparrow$ & Avg. SIoU$\uparrow$ & Min. SIoU$\uparrow$ \\
        \midrule
        \multicolumn{7}{l}{\textbf{Image inputs}} \\
        \midrule
        Codex (GPT-6) & 1.1238 & 3.6060 & 0.6148 & 0.8237 & 0.8813 & 0.8310 \\
        Articraft (GPT-6) & 1.3117 & 3.3719 & 0.5687 & 0.7615 & 0.8597 & 0.8068 \\
        \textbf{FACT} (GPT-6; original) & 1.0821 & \textbf{2.7884} & 0.6838 & 0.8391 & \textbf{0.9381} & \textbf{0.9166} \\
        \textbf{FACT} (GPT-6; parts) & \textbf{1.0317} & 2.9972 & \textbf{0.6937} & \textbf{0.8519} & 0.9317 & 0.9051 \\
        \hdashline
        \rowcolor{gray!12}
        Rodin (original) & 1.6840 & 1.9756 & 0.5663 & 0.7206 & 0.9393 & 0.9164 \\
        \rowcolor{gray!12}
        Rodin (parts) & 1.4353 & 2.6555 & 0.6041 & 0.7703 & 0.9336 & 0.9075 \\
        \midrule
        \multicolumn{7}{l}{\textbf{Oracle mesh inputs}} \\
        \midrule
        CAD-Recode & 1.2073 & 3.1952 & 0.6164 & 0.7990 & 0.8596 & 0.7475 \\
        MeshCoder & 3.4268 & 6.9213 & 0.3541 & 0.5018 & 0.6366 & 0.5035 \\
        PrimitiveAnything & 1.4028 & 4.7916 & 0.5354 & 0.7520 & 0.8241 & 0.7674 \\
        SuperFlex & 1.9266 & 3.2654 & 0.3937 & 0.6578 & 0.8852 & 0.8455 \\
        Codex (GPT-6) & 0.5402 & 1.0572 & 0.8783 & 0.9291 & 0.9888 & 0.9790 \\
        \textbf{FACT} (GPT-6)& \textbf{0.0874} & \textbf{0.3740} & \textbf{0.9899} & \textbf{0.9990} & \textbf{0.9902} & \textbf{0.9861} \\
        \bottomrule
    \end{tabular}%
\vspace{-6mm}
\end{table}

\section{Experiments}
\label{sec:experiment}

\subsection{Geometry Fidelity}
\label{sec:geometry_fidelity}

\paragraph{Experimental setup.}
We evaluate geometry reconstruction on 21 objects under image-input and oracle-mesh settings. In the image-input setting, we compare \textbf{FACT} against Codex and Articraft~\citep{codex,zhou2026articraft}. \textbf{FACT} reconstructs from either original or part meshes generated by Rodin Gen-2.5~\citep{zhang2024clay,zhang2025bang}. For oracle inputs, reference meshes replace generated geometry to isolate reconstruction quality from upstream 3D generation. We compare against CAD-Recode, MeshCoder, PrimitiveAnything, SuperFlex, and Codex~\citep{rukhovich2025cad,dai2026meshcoder,ye2025primitiveanything,tavernini2026superflex}. Learned baselines operate on whole-object geometry, whereas agentic methods can inspect the provided mesh structure. All agent-based methods use GPT-6 Astra~\citep{openai2026gpt6astra} with High reasoning level, with Codex using its default general-purpose workflow.

\paragraph{Evaluation protocol.}
Outputs are evaluated against the complete reference geometry in an aligned canonical pose. We report symmetric Chamfer distance (CD), the 95th-percentile reconstruction-to-reference surface distance (SD-P95), surface F-scores, and silhouette IoU (SIoU). CD and SD-P95 are normalized by the reference bounding-box diagonal and reported as percentages. Scores are averaged equally across objects.

\begin{figure}[t]
\vspace{-2mm}
    \centering
    \includegraphics[width=\linewidth]{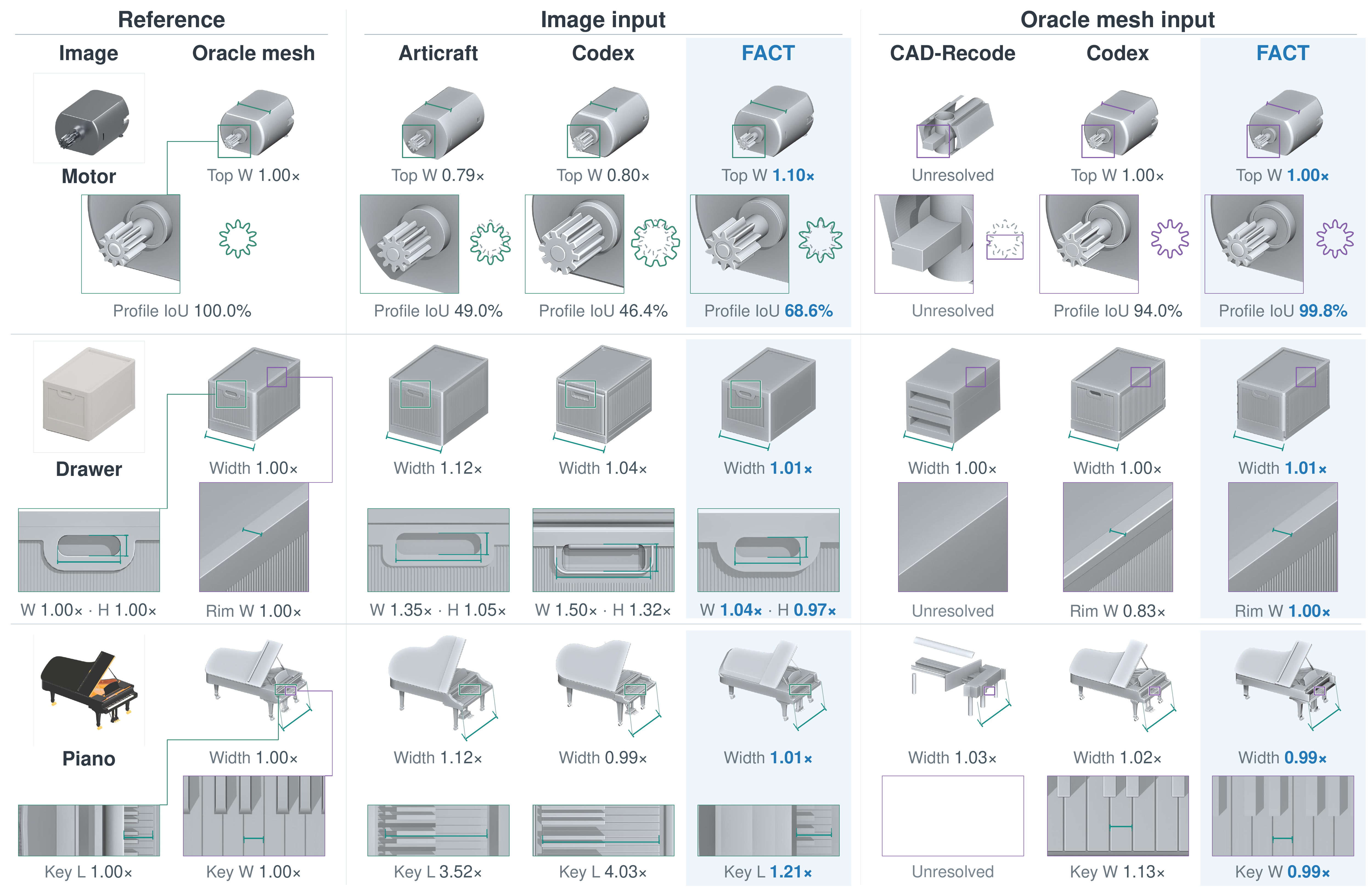}
    \vspace{-8mm}
    \caption{\textbf{Geometry fidelity.} \textbf{FACT} better preserves local structures and dimensions, while baselines can miss or distort fine details despite similar overall shapes.}
    \label{fig:geometry_fidelity}
    \vspace{-4mm}
\end{figure}

\begin{table}[t]
    \centering
    \vspace{-2mm}
    \caption{\textbf{Cumulative reconstruction ablation on oracle meshes.} Workflow stages are added sequentially to Codex (GPT-6); the final row is full \textbf{FACT}.}
    \label{tab:geometry_fidelity_ablation}
    \small
    \setlength{\tabcolsep}{3pt}
    \begin{tabular}{lcccccc}
        \toprule
        Condition & CD$\downarrow$ & SD-P95$\downarrow$ & F@1\%$\uparrow$ & F@2\%$\uparrow$ & Avg. SIoU$\uparrow$ & Min. SIoU$\uparrow$ \\
        \midrule
        Codex & 0.5402 & 1.0572 & 0.8783 & 0.9291 & 0.9888 & 0.9790 \\
        +Feature planning & 0.1414 & 0.6836 & 0.9730 & 0.9950 & 0.9903 & 0.9813 \\
        +Measurement & 0.0970 & 0.4878 & 0.9739 & 0.9926 & \textbf{0.9950} & \textbf{0.9915} \\
        +Refinement (\textbf{FACT}) & \textbf{0.0874} & \textbf{0.3740} & \textbf{0.9899} & \textbf{0.9990} & 0.9902 & 0.9861 \\
        \bottomrule
    \end{tabular}%
    \vspace{-4mm}
\end{table}

\paragraph{Reconstruction accuracy.}
With oracle inputs, \textbf{FACT} leads all six metrics and reduces CD by 83.8\% relative to Codex (Tab.~\ref{tab:geometry_fidelity_main}), supporting the value of structured reconstruction beyond access to accurate geometry alone. Both image-based variants also outperform Codex and Articraft, while improving CD and surface F-scores over their upstream meshes. Fig.~\ref{fig:geometry_fidelity} complements these aggregate results with matched-view comparisons and enlarged local structures. Fig.~\ref{fig:geometry_fidelity} further shows that \textbf{FACT}’s gains extend beyond overall shape agreement to more accurate local structures and dimensions.

\paragraph{Component analysis.}
The cumulative ablation in Tab.~\ref{tab:geometry_fidelity_ablation} shows progressive reductions in CD and SD-P95 as feature planning, targeted measurement, and diagnostic refinement are added. Refinement further reduces SD-P95 by 23.3\%, supporting the use of local feedback to correct residual surface discrepancies.

\subsection{Contact Fidelity}
\label{sec:agentic_coacd}

\paragraph{Experimental setup.}
We evaluate collision proxies on articulated objects, using shared reference geometry, rigid-body assignments, and target tasks. All methods use CoACD~\citep{wei2022coacd}. \textbf{Body} decomposes each rigid body as a whole, \textbf{Part} decomposes source parts within each body, and \textbf{Component} further separates closed connected components within each part. \textbf{DiagSearch} selects thresholds on the fixed Component partition using geometric, free-space, and motion diagnostics under hull-count budgets. \textbf{FACT} additionally revises local partitions and proxies using these diagnostics, followed by fidelity-constrained compression.

\begin{table}[t]
    \centering
    \caption{\textbf{Quantitative comparison of contact fidelity.} \textbf{FACT} achieves the lowest SCR and highest ISR with fewer hulls and lower physics-step time than DiagSearch. }
    \label{tab:agentic_coacd_main}
    \small
    \setlength{\tabcolsep}{3pt}
    \begin{tabular}{lccccccc} 
        \toprule
        Method & IoU (\%)$\uparrow$ & SCR (\%)$\downarrow$ & TFSO (\%)$\downarrow$ & ISR (\%)$\uparrow$ & Step time (ms)$\downarrow$ & $H\downarrow$ & $V\downarrow$ \\
        \midrule
        Body & 54.17 & 83.18 & 42.82 & 15.79 & 0.1035 & 76.1 & 9039.6 \\
        Part & 68.75 & 79.52 & 21.35 & 28.42 & 0.1013 & 223.6 & 22123.7 \\
        Component & 72.37 & 72.48 & 17.85 & 33.68 & 0.0956 & 306.2 & 21190.2 \\
        DiagSearch & 74.50 & 59.78 & 16.15 & 42.28 & 0.0988 & 326.0 & 22043.4 \\
        \textbf{FACT} & \textbf{79.43} & \textbf{0.00} & \textbf{6.79} & \textbf{99.47} & \textbf{0.0627} & 229.2 & 14493.7 \\
        \bottomrule
    \end{tabular}%
    \vspace{-3mm}
\end{table}

\begin{figure}[t]
    \centering
    \includegraphics[width=\linewidth]{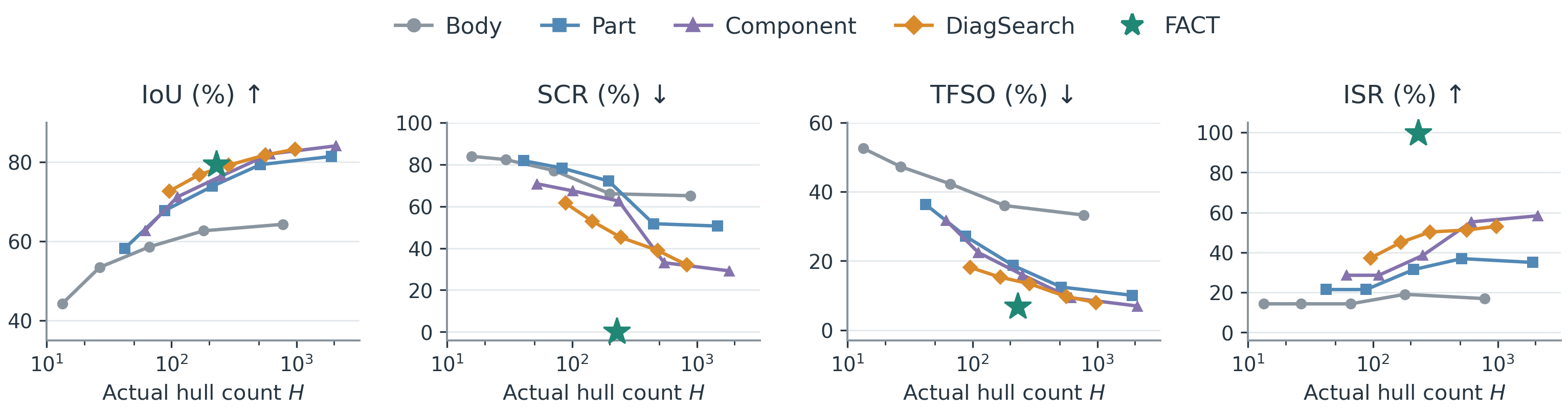}
    \vspace{-8mm}
    \caption{\textbf{Contact fidelity and proxy complexity.} Body, Part, and Component sweep CoACD thresholds; DiagSearch sweeps hull budgets. Failed settings are omitted; stars denote \textbf{FACT}.}
    \label{fig:agentic_coacd_line_plot}
\par\vspace{2mm}
    \centering
    \includegraphics[width=\linewidth]{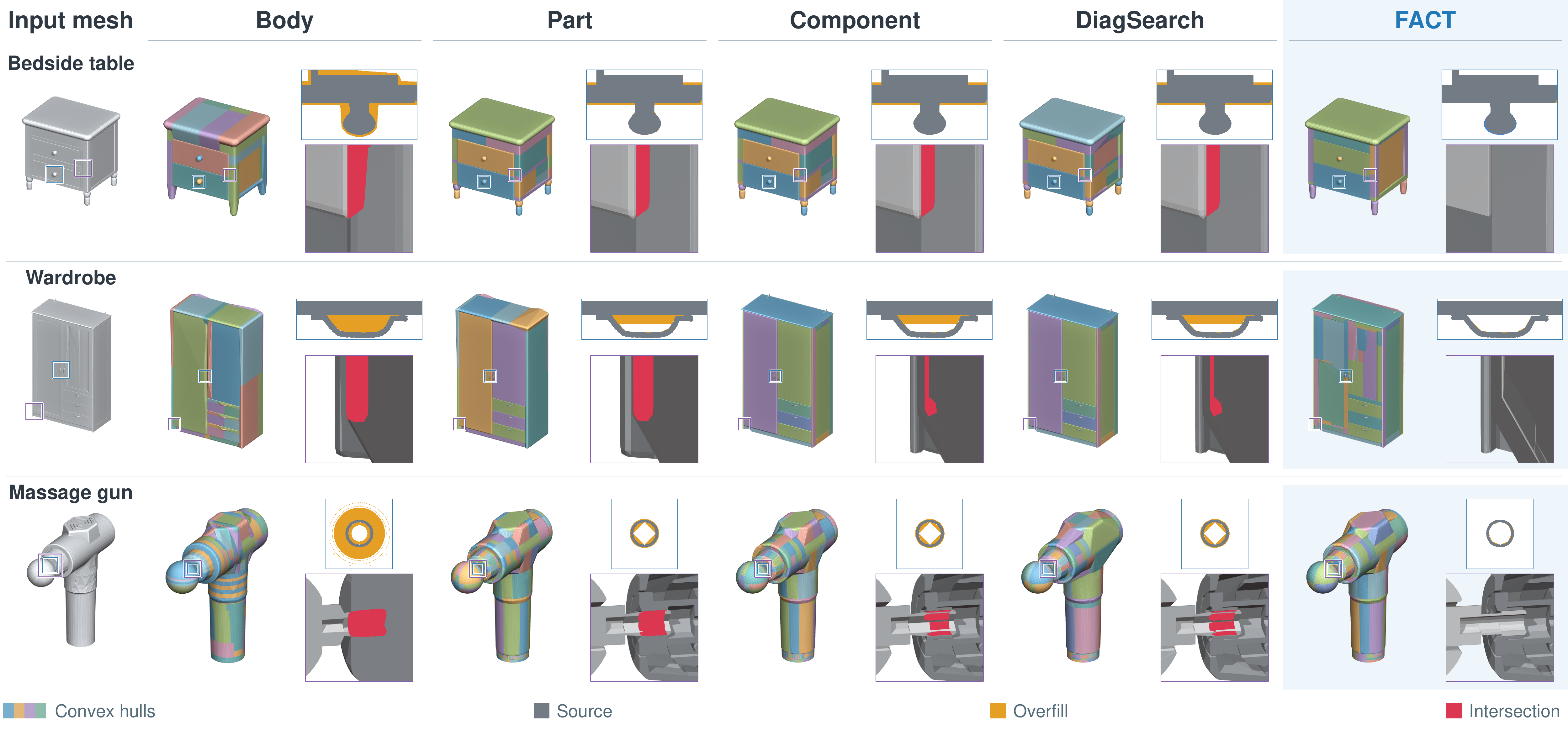}
    \vspace{-8mm}
    \caption{\textbf{Local collision proxy.} \textbf{FACT} preserves interaction-critical free space, including handle openings, while eliminating proxy-induced self-collisions.}
    \label{fig:agentic_coacd_visualization}
    \vspace{-6mm}
\end{figure}

\paragraph{Evaluation protocol.}
Each object is evaluated on one lifting or articulation task with ten gripper trials. Interaction success rate (ISR) is the percentage of trials satisfying the task goal and pass contact, penetration, and non-target-motion checks. We also report volumetric IoU, task free-space occupancy (TFSO), and self-collision rate (SCR) over the articulated motion space. Total hull and vertex counts (\(H,V\)) measure proxy complexity, while mean physics-step time measures simulation cost.

\paragraph{Interaction fidelity and efficiency.}
\textbf{FACT} achieves 99.47\% ISR versus 42.28\% for DiagSearch, with 29.7\% fewer hulls and 36.5\% lower physics-step time (Tab.~\ref{tab:agentic_coacd_main}). It also improves IoU and TFSO, with no self-collisions at the articulated motion space.

\paragraph{Fidelity–complexity trade-off.}
The sweeps in Fig.~\ref{fig:agentic_coacd_line_plot} show that increasing hull count generally improves volumetric fit, but does not ensure reliable interaction. \textbf{FACT} achieves 99.47\% ISR with 229.2 hulls on average, compared with 58.33\% ISR and 2,081.5 hulls for the finest evaluated Component configuration. These results support targeted repair followed by fidelity-constrained compression rather than finer decomposition alone. Fig.~\ref{fig:agentic_coacd_visualization} shows that \textbf{FACT} preserves the wardrobe’s handle opening and removes proxy-induced intersections.

\subsection{Dynamic Fidelity}
\label{sec:dynamic_fidelity}

\paragraph{Experimental setup.}
We evaluate dynamic prediction on held-out post-interaction clips of real objects. For each object, we first run the geometry and contact stages to obtain its reconstructed geometry \(G\), kinematic model \(K\), and refined collision proxies \(C\). Reference trajectories are extracted from these videos in the recovered kinematic model's joint coordinates, and object dimensions are measured with a ruler. We compare \textbf{FACT} against five direct-inference baselines (Tab.~\ref{tab:dynamic_fidelity}): GPT-5.6 Sol~\citep{openai2026gpt5.6sol}, GPT-6 Astra~\citep{openai2026gpt6astra}, Gemini-3.8-Flash~\citep{doshi2026gemini38flash}, Qwen3.8-Max~\citep{qwen38}, and Kimi-K3~\citep{team2026kimi}. These methods predict physical parameters from calibration videos or timestamped frames, whereas PhysX-Anything~\citep{cao2026physx} uses the initial frame. All predictions are instantiated in MuJoCo~\citep{todorov2012mujoco} with shared recovered geometry and joints, initial states, and solver settings. Response-model formulations follow each method's prediction procedure and may differ; the comparison therefore evaluates \textbf{FACT}'s full agentic dynamics calibration.

\begin{figure}[t]
    \centering
    \vspace{-2mm}
    \includegraphics[width=\linewidth]{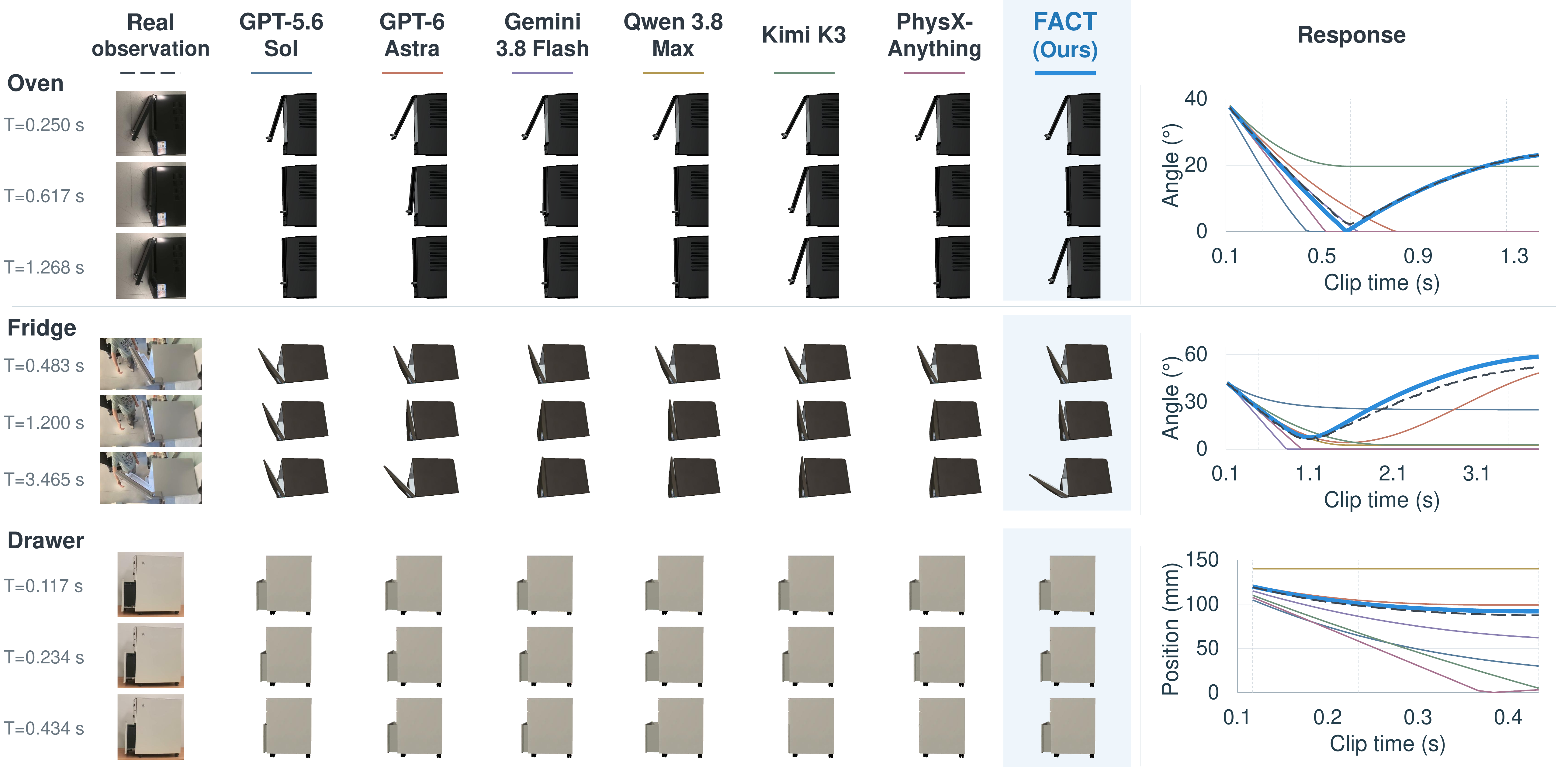}
    \vspace{-8mm}
    \caption{\textbf{Held-out motion prediction.} Time-aligned frames (left) and joint trajectories (right) compare real observations with simulated responses from \textbf{FACT} and the baselines for the oven, fridge, and drawer.}
    \label{fig:dynamic_fidelity}
    \vspace{-6mm}
\end{figure}

\paragraph{Evaluation protocol.}
Observed trajectories $q_t$ and simulated trajectories $\hat q_t$ are compared on a shared physical-time grid. Let $R=q_{\max}-q_{\min}$ denote the fixed joint range and $A=\max_t q_t-\min_t q_t$ the observed motion range. We report velocity
RMSE (VRMSE) normalized by $R$, trajectory MSE (TMSE) normalized by $A^2$, and the out-of-tolerance ratio OTR@10\%, the percentage of samples with absolute position error exceeding $0.1R$. Velocities are computed from smoothed trajectories by numerical differentiation, while TMSE and OTR use the unsmoothed trajectories. Evaluation windows exclude initialization and near-stationary tails.


\begin{table}[t]
    \centering
    \caption{\textbf{Dynamic prediction on held-out clips.}
    \textbf{FACT} outperforms direct parameter inference across all three motion metrics.}
    \label{tab:dynamic_fidelity}
    \small
    \setlength{\tabcolsep}{4pt}
    \begin{tabular}{lccc}
        \toprule
        Method
        & VRMSE$\downarrow$
        & TMSE$\downarrow$
        & OTR@10\%$\downarrow$ \\
        \midrule
        GPT-5.6 Sol      & 0.5181 & 0.6763 & 40.52 \\
        GPT-6 Astra      & 0.5853 & 1.1535 & 31.73 \\
        Gemini-3.8-Flash & 0.5566 & 0.1711 & 27.05 \\
        Qwen3.8-Max      & 0.6101 & 0.4236 & 36.49 \\
        Kimi-K3          & 0.8541 & 0.6995 & 46.97 \\
        PhysX-Anything   & 0.7165 & 0.7549 & 32.01 \\
        \textbf{FACT}    & \textbf{0.3786} & \textbf{0.0252} & \textbf{11.04} \\
        \bottomrule
    \end{tabular}%
\end{table}

\paragraph{Held-out prediction.}
\textbf{FACT} achieves the lowest error across all three metrics (Tab.~\ref{tab:dynamic_fidelity}), reducing TMSE by 85.3\% and OTR@10\% by 59.2\% relative to Gemini-3.8-Flash. Baseline performance varies across metrics, whereas our method consistently improves both velocity and positional agreement. Lower TMSE and OTR indicate that these gains extend beyond velocity matching to reproducing the trajectory on the shared physical-time grid. These results support the effectiveness of \textbf{FACT}'s agentic calibration loop. Fig.~\ref{fig:dynamic_fidelity} shows that \textbf{FACT} captures the motion reversals of the oven and fridge doors, whereas several baselines miss these transitions.

\section{Conclusion}
\label{sec:discussion}

We present \textbf{FACT}, an agentic framework for constructing articulated digital twins through quantitative measurements and simulation feedback. Its shared representation supports geometric reconstruction, task-aware collision refinement, and residual-guided dynamic calibration. Experiments show improved geometry, more reliable interaction with simpler proxies, and more accurate held-out motion prediction. \textbf{FACT} fixes geometry and kinematics during later refinement, limiting correction of upstream structural errors. Validation is restricted to selected interactions, sampled poses, and observed passive responses. Future work includes selective cross-stage revision and active acquisition of informative interactions.

\section*{AI Use Disclosure}
Generative AI tools were used to assist with writing and language polishing, and for literature retrieval and discovery. All AI-assisted content and references were reviewed and verified by the authors. The authors take full responsibility for the final content of this work.

\bibliography{iclr_conference}
\bibliographystyle{iclr_conference}

\newpage
\appendix
\section{Data and Experimental Settings}
\label{app:data-settings}

Geometry evaluation (Tab.~\ref{tab:geometry_fidelity_main}--\ref{tab:geometry_fidelity_ablation}) uses all 21 reference assets. Contact evaluation (Tab.~\ref{tab:agentic_coacd_main}) uses a 19-object subset; \texttt{grand\_piano} and
\texttt{oven\_6\_333} are excluded because extensive non-manifold geometry prevents reliable construction of the closed-material references required by the contact metrics. The dynamics evaluation (Tab.~\ref{tab:dynamic_fidelity}) uses six separate real objects. Each real object is first reconstructed through the geometry and contact stages, and the resulting articulated twin is then used for dynamics calibration and evaluation.

\begin{wraptable}{r}{0.43\linewidth}
\vspace{-5em}
\centering
\caption{Reference dataset for the geometry and contact stages. V: ArtVIP; A: commercial artist asset.}
\label{tab:app-assets}
\scriptsize
\setlength{\tabcolsep}{6pt}
\begin{tabular}{lcc}
\toprule
Asset ID & Source & Kinematic bodies \\
\midrule
bakingchamber             & V & 2 \\
bedsidetable              & V & 3 \\
brimnes\_cabinet          & V & 3 \\
cabinet\_1                & V & 3 \\
drawer\_kit\_3            & V & 6 \\
grand\_piano              & A & 8 \\
handheld\_vacuum\_cleaner & A & 1 \\
humidifier                & A & 1 \\
lipstick                  & A & 2 \\
massage\_gun              & A & 2 \\
match                     & V & 2 \\
microwave\_1              & V & 3 \\
motor                     & A & 2 \\
musken\_wardrobe          & V & 6 \\
oven\_6\_333              & V & 6 \\
pet\_feeder               & A & 1 \\
power\_bank               & A & 1 \\
pressure\_pump\_2         & V & 2 \\
refrigerator\_8           & V & 7 \\
rolling\_washer\_5        & V & 3 \\
washing\_machine\_2       & V & 4 \\
\bottomrule
\end{tabular}
\vspace{-5em}
\end{wraptable}

\subsection{Reference assets}
\label{app:reference-assets}

The reference collection contains 13 ArtVIP~\citep{jin2026artvip} assets and eight artist-created commercial assets. Tab.~\ref{tab:app-assets} lists the assets and their reviewed articulation structures. Geometry evaluation uses all 21 assets and the complete reference surfaces, including interior geometry. Contact evaluation uses 19 assets and reviewed closed material units, excluding geometry marked as visual-only. Objects with a single kinematic body are excluded from SCR, for which self-collision is not applicable. Repairs to open boundaries and material solids, as well as reviewed body assignments and joints, are shared across all contact methods.

\subsection{Real-object recordings}
\label{app:real-recordings}

Tab.~\ref{tab:app-recordings} summarizes the six real objects and their recording coverage. The reported clip count is the total number of recorded clips for each object. For each object, one or two clips are held out exclusively for evaluation, while all remaining clips are available for observation extraction, response-model selection, parameter fitting, and stopping decisions. Recordings are $1280\times720$ at approximately 60 FPS. A measured exterior dimension determines the scale factor $s=L_{\mathrm{real}}/L_{\mathrm{mesh}}$. The resulting uniform scale is applied consistently to visual and collision geometry, joint locations, and prismatic travel, while revolute joint angles remain unchanged.

\begin{table}[htbp]
\centering
\caption{Real-object inventory and recording coverage.}
\label{tab:app-recordings}
\small
\setlength{\tabcolsep}{4pt}
\begin{tabular}{lp{3.2cm}crr}
\toprule
Object & Observed motion & Kinematic bodies & Duration (s) & Clips \\
\midrule
Chest    & Lid with passive stays   & 4 & 25.84 & 2 \\
Drawer   & Lower drawer             & 4 & 21.93 & 6 \\
Trashcan & Coupled lid and pedal    & 3 & 18.91 & 2 \\
Cabinet  & Middle door              & 4 & 17.08 & 3 \\
Fridge   & Upper door               & 4 & 41.70 & 7 \\
Oven     & Vertical-axis door       & 2 & 32.47 & 9 \\
\bottomrule
\end{tabular}
\end{table}
\section{Geometry Fidelity}
\label{app:geometry}

\subsection{Implementation and validation}

\paragraph{Feature implementation.}
Feature plans associate editable features with source regions while preserving node identities. After applying scene-node transforms, GLB/glTF coordinates are mapped to Blender as $(x,y,z)\mapsto(x,-z,y)$; OBJ/STL coordinates are unchanged. Coordinate frames are defined from observed geometric and semantic datums. Parameter records track measurements, derived constraints, agent adjustments, defaults, and user inputs together with their source regions, frames, and measurement settings. Contour reduction and section sampling are selected per feature type. Reconstructed meshes are generated from feature parameters rather than copied source triangles, and exported part and feature identifiers preserve assembly correspondences.

\paragraph{Validation.}
Let $D_M$ denote the source bounding-box diagonal. Automated validation requires per-axis dimension errors below 5\%, center error below $0.02D_M$, both directed global P95 surface distances below $0.02D_M$, and canonical-view silhouette IoU of at least 0.90. All checks are performed in source assembly coordinates without alignment. When source-part correspondence is available, part-level checks use 2,048 samples per direction and a threshold of $0.02D_k$, where $D_k$ is the source-part diagonal. For unsegmented inputs, recovered regions are used only for visual diagnosis and are not treated as source-part ground truth. Numerical validation is supplemented by structural inspection, with intentional simplifications recorded as residuals.

\begin{table}[htbp]
\centering
\caption{Inference configurations for learned geometry baselines.}
\label{tab:geometry-baseline-configurations}
\small
\setlength{\tabcolsep}{4pt}
\begin{tabular}{@{}lp{10.2cm}@{}}
\toprule
Method & Configuration \\
\midrule
CAD-Recode & 256 input points; 10 generated programs; CD-based selection among executable candidates \\
MeshCoder & 16,384 normalized input points; generated Blender code executed without manual repair \\
PrimitiveAnything & 10,000 surface points with normals sampled from the normalized input mesh; official autoregressive inference pipeline \\
SuperFlex & 4,096-point input point cloud; feed-forward deformable-superquadric checkpoint; no object-specific optimization \\
\bottomrule
\end{tabular}
\end{table}

\subsection{Inputs and baseline configurations}

For the image-input setting, Codex and Articraft receive only the input image, while FACT incorporates Rodin generation into its pipeline, producing an original or part-split asset that is subsequently frozen for reconstruction. For oracle-mesh evaluation, learned baselines receive the whole-object geometry without part grouping, while agentic methods retain the scene hierarchy and may inspect individual parts. All agentic geometry methods use GPT-6 Astra with High reasoning level. Codex operates through Blender MCP without the FACT reconstruction workflow, while Articraft follows its native pipeline. Inference configurations for the learned baselines are summarized in Tab.~\ref{tab:geometry-baseline-configurations}.

\subsection{Geometry evaluation metrics}
\label{benchmark-metric-implementation}

\paragraph{Pose and alignment.}
All metrics are computed on whole-object visual surfaces in a canonical pose. Articraft URDFs apply the visual-mesh scales and link/joint transforms, while FACT assets use their closed or default pose. Let $c_P,c_S$ denote the AABB centers of the candidate and reference, and let $\ell_P,\ell_S$ denote their longest AABB side lengths. The candidate is aligned by the uniform transform
\[
p' = c_S + \frac{\ell_S}{\ell_P}(p-c_P).
\]
No metric-driven rotation search, ICP, nonuniform scaling, or per-part alignment is applied. The resulting scores therefore measure normalized shape and relative assembly dimensions, rather than absolute scale or global translation.

\paragraph{Surface distances and F-scores.}
Let $S$ denote the reference surface, $P$ the aligned candidate surface, and $D_S$ the diagonal of the reference AABB. We draw $n=4{,}096$ area-weighted surface samples in each direction, excluding nonfinite triangles and triangles with area at most $10^{-15}$. Distances are unsquared Euclidean point-to-surface distances. For $x_i\sim S$ and $y_j\sim P$, define
\[
a_i=\frac{\min_{p\in P}\|x_i-p\|_2}{D_S},
\qquad
b_j=\frac{\min_{s\in S}\|y_j-s\|_2}{D_S}.
\]
We report
\[
\mathrm{CD}=\frac{100}{2n}\left(\sum_i a_i+\sum_j b_j\right),
\qquad
\mathrm{SD\text{-}P95}=100\,Q_{0.95}(b_1,\ldots,b_n),
\]
where $Q_{0.95}$ uses linear percentile interpolation. SD-P95 measures reconstruction-to-reference error only; missing reference geometry instead contributes to the reverse CD term and F-score recall. For $\tau\in\{0.01,0.02\}$,
\[
p_\tau=\frac{1}{n}\sum_j\mathbf{1}[b_j\le\tau],
\qquad
r_\tau=\frac{1}{n}\sum_i\mathbf{1}[a_i\le\tau],
\qquad
F_\tau=\frac{2p_\tau r_\tau}{p_\tau+r_\tau},
\]
with $F_\tau=0$ when $p_\tau+r_\tau=0$. F@1\% and F@2\% are reported on $[0,1]$.

\paragraph{Silhouette IoU.}
We rasterize triangle projections into binary $512\times512$ masks using Pillow without antialiasing. The three orthographic projections are $(-x,z)$, $(y,z)$, and $(x,y)$. Each candidate--reference pair shares the union of their projected bounds, a common isotropic pixel scale, and a 31-pixel margin. Avg. SIoU and Min. SIoU denote the mean and minimum IoU across the three views, respectively.

\subsection{Case Study: Baking-chamber Handle Refinement}
The baking chamber contains eight parts represented by 15 generated meshes. In source coordinates, the agent extracts a handle section from the 19 unique vertices satisfying $|z+0.240055|<10^{-5}$. Section extrema determine the center $c_x$, half-width $h_x$, front coordinate $y_f$, and depth $d$. The agent selects a half-superellipse parameterization,
\[
u_i=\frac{|x_i-c_x|}{h_x},
\qquad
v_i=\frac{y_i-y_f}{d},
\]
and the numerical tool fits the superellipse exponent as
\[
e^\star=\arg\min_{1.2\le e\le5}\sum_i\left(u_i^e+v_i^e-1\right)^2.
\]
The resulting parameters are $c_x\approx0.02538$, $h_x\approx0.06353$, $y_f\approx-0.48697$, $d\approx0.06631$, and $e^\star\approx2.6593$. The agent then revises the feature representation to loft height-varying half-superellipse sections while preserving the observed open mating face.

Although the initial reconstruction passed whole-object validation, part-level diagnostics revealed an inaccurate handle profile and spurious closure surfaces (Tab.~\ref{tab:geometry-local-refinement}). The agent therefore remeasures the handle and revises the affected feature representations. After revision, all eight parts satisfy the local threshold, despite a small increase in the global reconstruction-to-source P95 distance.

\begin{table}[htbp]
\centering
\caption{Local baking-chamber refinement. Surface distances are the maximum of the two directed part P95 values, divided by $D_k$.}
\label{tab:geometry-local-refinement}
\small
\setlength{\tabcolsep}{4pt}
\begin{tabular}{@{}lrrp{5.6cm}@{}}
\toprule
Part & Before & After & Revision \\
\midrule
Handle trim & 0.037068 & 0.009469 & Remove rear closure \\
Handle & 0.087434 & 0.009444 & Remeasure; replace extrusion with open loft \\
Dark surround & 0.114964 & 0.002874 & Remove concave mating surface \\
\bottomrule
\end{tabular}
\end{table}
\section{Contact Fidelity}
\label{app:contact}
\label{contact-fidelity-constraints-tasks-and-evaluation-protocol}

\subsection{Proxy Initialization, Repair, and Compression}
\label{app:contact-search}

\paragraph{Proxy Initialization.}
Within each rigid body, the reference material is the union of reviewed solids, with overlapping volume counted once. Repartitioning preserves source-face ownership and never crosses body boundaries. The initial proxy $C^0$ is the Component decomposition: closed, consistently oriented face-connected components are separated within each source part, while residual components remain grouped. The object-level hull budget is fixed as $H_{\max}=H(C^0)$.

CoACD 1.0.14 uses threshold 0.05, automatic preprocessing, preprocessing resolution 50, resolution 2000, and MCTS nodes/iterations/depth of 20/150/3. We retain the default merging behavior, disable PCA, decimation, and extrusion, use the 256-vertex setting and convex-hull approximation, and impose no per-call hull cap. Body, Part, and Component use the default threshold; DiagSearch uses its frozen selection under a 512-hull budget, whereas FACT uses the object-specific budget $H_{\max}$.

\paragraph{Repair search.}
The agent proposes local repartitions and regional decomposition settings, while the numerical backend regenerates and evaluates the affected proxies. A candidate is feasible only if it satisfies the hull budget $H(C)\le H_{\max}$, passes the sampled motion checks, and keeps the prescribed volumetric and free-space errors within their allowed increases relative to $C^0$. Candidates violating any of these constraints are rejected. Among feasible candidates, the search first minimizes worst-region contact P95; candidates within $0.0005D$ of the best value are compared by development loss, and those within 0.005 of the best development loss are finally ranked by proxy complexity $(H,V)$, where $D$ is the contact-reference diagonal. The development loss equally averages $1-\mathrm{IoU}$, ROI occupancy, and motion error, with the motion term omitted for static objects. The selected feasible proxy is fixed as the repair anchor $C^a$.

\paragraph{Compression.}
All compression candidates are evaluated against the fixed anchor $C^a$, rather than the preceding iterate. For each body, the allowed increases in $(1-\mathrm{IoU},m,x)$ are $(0.02,0.002,0.05)$, and each eligible development ROI permits an occupancy increase of at most 0.01. Auxiliary source-normal guards allow P95 and maximum displacement to increase by at most $0.0005D$ and $0.002D$, respectively, with no increase in missing normal intersections.

\subsection{Contact Diagnostics and Metrics}
\label{app:contact-metrics}

Let $D$ denote the reference object's bounding-box diagonal.

\paragraph{Volumetric fidelity.}
For reference material $G_b$ and proxy union $C_b$ of body $b$, we measure
\[
\mathrm{IoU}_b=\frac{|G_b\cap C_b|}{|G_b\cup C_b|},\qquad
m_b=\frac{|G_b\setminus C_b|}{|G_b|},\qquad
x_b=\frac{|C_b\setminus G_b|}{|G_b|},
\]
where $m_b$ and $x_b$ are the missing- and excess-volume ratios, respectively. Per-object values are averaged equally across bodies. Volume estimation uses eight scrambled Sobol batches with $2^{15}$ development or $2^{17}$ evaluation queries. Samples combine uniform sampling in a source-derived domain, initially padded by $0.1D$, with volume-proportional sampling from reference-material PCA boxes; inverse-density weights correct for overlapping boxes. All methods share the same samples and reference labels, which are recomputed if the sampling domain expands. Reference occupancy is determined by three signed-ray tests with tolerance $10^{-9}D$. Ambiguous queries define lower and upper IoU bounds, and we report their midpoint. Missing and excess ratios use only queries with known reference labels; uncertainty from ambiguous queries and across Sobol batches is tracked separately.

\paragraph{Self-collision.}
For each sampled motion path, let $Q^{\rm free}$ contain poses at which the reference bodies are classified as collision-free. We compute
\[
\mathrm{SCR}
=\frac{100}{|Q^{\rm free}|}
\sum_{q\in Q^{\rm free}} I_\epsilon(C,q),
\]
where $I_\epsilon(C,q)=1$ if an interbody hull intersection contains a ball of radius $\epsilon=10^{-4}D$, tested using inward-offset halfspaces. Same-body overlaps are ignored. Evaluation poses are densely sampled throughout the prescribed articulated motion space of each object. Rates are averaged equally over eligible motion paths, available decomposition seeds, and the 15 articulated contact objects.

\paragraph{Task free-space occupancy.}
For region $r$ at pose $q$ with source-defined probe distribution $\mu_{r,q}$, we measure
\[
\mathrm{TFSO}_{r,q}
=100\,\Pr_{p\sim\mu_{r,q}}
\left[
p\in\bigcup_b T_b(q)C_b
\,\middle|\,
p\text{ is confirmed reference-free}
\right],
\]
where $T_b(q)$ denotes the transform of body $b$. Free probes are determined from the full reference-material union independently of the candidate proxy. Access probes are sampled from source surfaces by area with offsets in $[10^{-4}D,0.02D]$, while motion-clearance probes use $0.01D$ bands near neighboring bodies. A region is eligible only with at least 128 confirmed-free samples overall and 16 per batch. Scores are averaged equally across batches, eligible poses, regions within each access or clearance category, and available categories.

\subsection{Interaction protocol}
\label{app:contact-interaction}

Each object is assigned one prescribed interaction task, executed by an actuated Cartesian gripper with wrist rotation and opposing fingers or by a pressing pad. Object joints remain passive and move only through physical contact. Task trajectories use cubic smoothstep interpolation, $3u^2-2u^3$, with normalized time $u\in[0,1]$. Controller parameters and task durations are fixed across methods. Each proxy is evaluated over ten trials: trial 00 uses the nominal setup, while trials 01--09 apply uniformly sampled position offsets with half-ranges $(1,0.4,0.4)$ mm. The same offsets are used across methods, with no orientation perturbation.

\paragraph{Success criteria.}
Articulation tasks must maintain at least 90\% of the target progress throughout the final hold. Static lifting tasks require a center-of-mass rise of at least 90 mm and a source-to-floor clearance of at least 20 mm throughout the sampled hold states. All trials additionally require penetration of at most 1 mm and tracking error of at most 20 mm; where applicable, non-target joint displacement must remain below 0.03 m for prismatic joints or 0.03 rad for revolute joints. Forbidden non-target contacts and solver warnings constitute failure. All trials run for the prescribed horizon, and failure to reach or maintain the task goal is counted as unsuccessful.

\subsection{Simulation and timing}
\label{app:contact-simulation}

Contact evaluation uses MuJoCo 3.13.0 with a 1 ms timestep, \texttt{implicitfast} integration, a Newton solver with 80 iterations, an elliptic friction cone, and gravity $(0,0,-9.81)$ m/s$^2$. The solver tolerance is $10^{-9}$, except $10^{-10}$ for the cabinet. Contact parameters are \nolinkurl{solref=(0.002,1)}, \nolinkurl{solimp=(0.95,0.99,0.001)}, \texttt{condim=4}, zero margin, and friction $(0.8,0.005,0.0001)$. Within each object, body inertials, joint properties, contact masks, and controllers are fixed across methods.

Physics-step timing is measured on an Intel Core i9-14900K using one serial pass per proxy. Native \texttt{mj\_step} calls are timed after a 0.5 s warm-up; control, rendering, model loading, and source-geometry audits are excluded. Reported step times are averaged over simulation steps, decomposition seeds, and objects.

\subsection{Microwave refinement trace}
\label{app:contact-microwave}

The initial shell proxy induces false collisions with the door and turntable. The agent repartitions the shell along measured cavity planes into six structural regions. It then sets the CoACD thresholds for the door frame and shell to 0.01 and 0.025, respectively, with preprocessing disabled. A 64-vertex turntable candidate passes the sampled motion checks but is rejected for 45.23\% missing material; restoring the thin-disc boundary with 100 vertices resolves this failure. Tab.~\ref{tab:app-contact-trace} reports the corresponding development and final measurements.

Starting from the repaired anchor $C^a$, compression removes 82 hulls. The largest per-body increases in $(1-\mathrm{IoU},m,x)$ are approximately $(0.003254,0.001529,0.001767)$, and the largest ROI occupancy increase is 0.005859, all within the prescribed fidelity bounds. Subsequent local edits either violate these guards or fail to further reduce the hull count, yielding a local stopping point. The final proxy succeeds in all ten interaction trials while retaining bounded approximation error.

\begin{table}[htbp]
\centering
\caption{Microwave refinement trace from the initial proxy $C^0$ to the repaired anchor $C^a$ and compressed proxy $C^{\rm final}$. IoU and ROI occupancy are percentages; fidelity constraints are enforced per body and region rather than on aggregate values. False/free reports proxy-induced collision poses among reference-free samples.}
\label{tab:app-contact-trace}
\small
\setlength{\tabcolsep}{4pt}
\begin{tabular}{lccccc}
\toprule
Stage / samples & $H\downarrow$ & $V\downarrow$ & IoU $\uparrow$ & ROI occ. $\downarrow$ & False/free $\downarrow$ \\
\midrule
$C^0$ / development & 420 & 19,641 & 95.6736 & 24.4141 & 93/93 \\
$C^a$ / development & 364 & 14,449 & 99.7027 & 16.9789 & 0/93 \\
$C^{\rm final}$ / development & 282 & 12,385 & 99.5943 & 17.0410 & 0/93 \\
$C^{\rm final}$ / final & 282 & 12,385 & 99.5784 & 16.9522 & 0/603 \\
\bottomrule
\end{tabular}
\end{table}

\section{Dynamic Fidelity}
\label{app:dynamics}

We evaluate passive-response prediction on six reconstructed real objects. For each object, one or two clips are reserved exclusively for evaluation, while all remaining clips are used for calibration. The same split is used for FACT and all baselines: model construction or parameter inference uses only the calibration clips, and the held-out clips are used only for final evaluation (Section~\ref{app:real-recordings}).

\subsection{Trajectory extraction}
\label{app:dynamics-observation}

Timestamped videos are converted to joint-coordinate trajectories in the reconstructed kinematic model. We first identify hand-free intervals and define release as the first reliable passive frame. Object motion is recovered from image-space features, including tracked points, silhouettes, and edges, and mapped to the corresponding joint coordinates using geometry-aware planar mappings or pivot constraints. Measurements are filtered by detector confidence and local temporal consistency; isolated gaps are interpolated only when supported by reliable neighboring observations, while unresolved detections are masked. Evaluation uses the accepted, unsmoothed position measurements. All methods share the same trajectory references, coordinate mappings, masks, and evaluation windows. 

\subsection{Response models}
\label{mechanism-specific-response-models}

Let $q$ denote the joint coordinate defined by the reconstructed kinematic model, oriented positively toward opening, and let $v=\dot q$. From the reconstructed mechanism and observed passive motion, the agent selects a response-model family, diagnoses systematic residuals, and revises the model when needed. Numerical tools then fit the free physical parameters of the selected model. Geometry, kinematics, collision proxies, inertial estimates, and fixed boundary conditions remain unchanged during each fit.

Tab.~\ref{tab:app-response-models} summarizes the final agent-constructed response models. These models capture the dominant passive effects observed across the six mechanisms, including dry and viscous resistance, state-dependent restoring terms, piecewise spring--damper behavior, and impact-like closure responses.

\begin{table}[htbp]
\centering
\caption{Agent-constructed response models for dynamics calibration. Reported values are the final fitted parameters.}
\label{tab:app-response-models}
\small
\renewcommand{\arraystretch}{1.12}
\setlength{\tabcolsep}{5pt}
\begin{tabular}{@{}lp{0.82\linewidth}@{}}
\toprule
Object & Response model and fitted parameters \\
\midrule

Chest &
Gravity-driven linkage with Coulomb lid resistance,
$\tau_{\mathrm{res}}=-\tau_c\operatorname{Sign}(v)$.\\
& \textit{Fit:} $\tau_c=0.56757~\mathrm{N\,m}$. \\
\addlinespace[2pt]

Drawer &
Viscous and Coulomb resistance,
$\ddot q=-\beta v-\alpha\operatorname{Sign}(v)$.\\
& \textit{Fit:} $\beta=4.04196~\mathrm{s}^{-1}$,
$\alpha=0.375197~\mathrm{m/s^2}$. \\
\addlinespace[2pt]

Trashcan &
Coupled lid--pedal mechanism with viscous lid resistance,
$\tau_{\mathrm{res}}=-bv$.\\
& \textit{Fit:} $b=0.00790463~\mathrm{N\,m\,s/rad}$. \\
\addlinespace[2pt]

Cabinet &
Piecewise spring--damper response with distinct closing,
intermediate, and opening regimes.\\
& \textit{Fit:} $(k_c,b_c)=(0.19321,0.34609)$,
$(k_o,b_o)=(11.4396,0.505714)$,
$q_o=89.8416^\circ$. \\
\addlinespace[2pt]

Fridge &
State-dependent restoring response with smooth velocity resistance,
$\ddot q=Ae^{-q/w}-f\tanh(v/0.025)$.\\
& \textit{Fit:} $A=21.8265$,
$w=0.064639$,
$f=0.212000$. \\
\addlinespace[2pt]

Oven &
Coulomb deceleration with state-triggered closure restitution,
$\ddot q=-a\operatorname{Sign}(v)$.\\
& \textit{Fit:} $a=0.842825~\mathrm{rad/s^2}$,
$e=0.734856$. \\

\bottomrule
\end{tabular}
\end{table}
For the cabinet, the final piecewise response is
\[
\tau(q,v)=
\begin{cases}
-k_cq-b_cv, & q<q_1,\\
-b_mv, & q_1\le q\le q_2,\\
k_o(q_o-q)-b_ov, & q>q_2,
\end{cases}
\]
where the transition locations and intermediate damping are fixed during parameter fitting. For the fridge, the fitted exponential restoring term is combined with a fixed state-dependent gate during replay. For the oven, reaching the closure boundary triggers a restitution update determined by the fitted coefficient $e$. These fixed transition and boundary rules define the response family selected by the agent but are not themselves optimized in the final fit.

\subsection{Parameter fitting}
\label{the-executed-calibration-problems}

We instantiate the objective in Eq.~(2) as a normalized weighted
least-squares problem. For accepted observations $\mathcal T_k$, shared
physical parameters $\theta$, and sequence-specific initial states
$\xi_k=(q_{0,k},v_{0,k})$, we solve
\[
\min_{(\theta,\{\xi_k\})\in\Omega}
\frac12\sum_k
\left[
c_k\sum_{i\in\mathcal T_k}
\left(
\frac{q_k^{\rm sim}(t_i;\theta,\xi_k)-q_{ki}^{\rm obs}}{s_q}
\right)^2
+R_k(\xi_k)
\right].
\]
Here, $s_q$ normalizes trajectory residuals, $c_k$ controls sequence
weighting, and $R_k$ optionally regularizes uncertain initial states.
Geometry, kinematics, inertial estimates, camera mappings, and time
offsets remain fixed during fitting.

\begin{table}[htbp]
\centering
\caption{Candidate parameter and initial-state bounds considered during calibration and model revision. Final selected models may retain only a subset of the listed parameters.}
\label{tab:app-fit-settings}
\small
\renewcommand{\arraystretch}{1.12}
\setlength{\tabcolsep}{5pt}
\begin{tabular}{@{}l
>{\raggedright\arraybackslash}p{0.52\linewidth}
>{\raggedright\arraybackslash}p{0.34\linewidth}@{}}
\toprule
Object & Parameter bounds & Initial-state bounds \\
\midrule

Chest &
$\tau_c\in[0.05,1.5]$, $b\in[0,0.12]$. &
$q_0:\pm1.3^\circ$;
$v_0:\pm r_v$, with $v_0\le0$. \\

\addlinespace[2pt]

Drawer &
$\beta\in[0.01,30]$, $\alpha\in[0,10]$. &
$q_0:\pm0.004\,\mathrm{m}$;
$v_0:\pm\max\!\left(0.3\,\mathrm{m/s},
0.35|\widetilde v_0|\right)$. \\

\addlinespace[2pt]

Trashcan &
$b\in[0,0.04]$, $\tau_c\in[0,0.035]$. &
$q_0:\pm3^\circ$;
$v_0:\pm57.3^\circ/\mathrm{s}$, with $v_0\le0$. \\

\addlinespace[2pt]

Cabinet &
$k_c\in[0.001,2]$, $b_c\in[0.001,3]$,
$k_o\in[0.1,40]$, $b_o\in[0.005,3]$,
$q_o\in[86^\circ,97^\circ]$. &
$q_0:\pm2^\circ$;
$v_0:\pm20^\circ/\mathrm{s}$. \\

\addlinespace[2pt]

Fridge &
$A\in[0.1,200]$, $w\in[0.015,0.25]$,
$f\in[0.01,1.5]$. &
$q_0:\pm1.5^\circ$;
$v_0:\pm8^\circ/\mathrm{s}$. \\

\addlinespace[2pt]

Oven &
$a\in[0.03,8]$, $b\in[0,8]$,
$e\in[0.05,0.95]$, $\delta\in[0,4^\circ]$. &
$q_0:\pm2^\circ$, with $q_0\ge0$;
$v_0:\pm\max\!\left(
20^\circ/\mathrm{s},
0.25|\widetilde v_0|
\right)$. \\

\bottomrule
\end{tabular}
\end{table}

Initial states are estimated from short release windows using low-order polynomial fits and, when uncertain, optimized within bounded neighborhoods of these estimates. The numerical backend uses bounded nonlinear least squares with model-specific forward evaluators: direct simulation when linkage dynamics are retained, analytical responses when available, and numerical integration otherwise. Multiple initializations are evaluated when needed, and the lowest-cost feasible solution is retained. Tab.~\ref{tab:app-fit-settings} reports the parameter bounds considered during calibration and model revision, together with the initial-state bounds. The final selected response model may retain only a subset of these candidate parameters.

\paragraph{Residual-guided model revision.}
After each fit, the agent inspects trajectory residuals to determine whether errors arise from parameter values, uncertain initial conditions, or an inadequate response-model form. It then revises the optimized variables, bounds, or model structure and refits numerically. When multiple response models explain the observations, we retain the simplest model satisfying the prescribed per-sequence error criteria.

\subsection{Replay and evaluation}
\label{app:dynamics-replay}

\paragraph{Baseline inputs.}
All direct-inference baselines operate only on the calibration observations. Language-model baselines receive calibration motion videos or timestamped frames in the modality supported by their respective interfaces, while PhysX-Anything receives an initial frame from the calibration recordings. Their inferred response models or physical parameters are then fixed and replayed on the held-out evaluation clips. No baseline receives the evaluation trajectories during model or parameter inference. Because the baselines differ in input modality and response-model parameterization, the comparison evaluates end-to-end motion prediction rather than enforcing identical intermediate representations.

\paragraph{Replay protocol.}
All methods are evaluated using the same reconstructed geometry, joint coordinates, observation-derived initial states, and numerical settings within each object. Initial states are estimated from a short release window and are not fitted to the evaluation trajectory. Method-specific response models and inferred physical parameters are otherwise preserved. Forward replay uses MuJoCo 3.13.0 with Euler integration, Newton solving, a dense Jacobian, elliptic friction cones, tolerance $10^{-10}$, gravity $(0,0,-9.81)$ m/s$^2$, and zero external input after release.

\paragraph{Temporal alignment and scoring windows.}
Observed and simulated trajectories are linearly interpolated onto a common physical-time grid at approximately 60 Hz. Position metrics use the accepted unsmoothed observations, while velocities are obtained using a shared smoothing and numerical-differentiation procedure. Missing or unreliable observations are masked consistently. Evaluation windows and stationary-tail truncation are determined from the reference trajectory alone and then shared across methods.

\paragraph{Motion metrics.}
Let $\mathcal V_q$ and $\mathcal V_v$ denote the valid position and velocity indices, $R=q_{\max}-q_{\min}$ the fixed joint range, and
\[
A=\max_{i\in\mathcal V_q}q_i-\min_{i\in\mathcal V_q}q_i
\]
the observed motion amplitude. We report
\[
\begin{aligned}
\mathrm{VRMSE}
&=
\frac{1}{R}
\sqrt{
\frac{1}{|\mathcal V_v|}
\sum_{i\in\mathcal V_v}
(\widehat v_i-v_i)^2
},\\
\mathrm{TMSE}
&=
\frac{
\sum_{i\in\mathcal V_q}
(\widehat q_i-q_i)^2
}{
|\mathcal V_q|A^2
},\\
\text{OTR@10\%}
&=
\frac{100}{|\mathcal V_q|}
\sum_{i\in\mathcal V_q}
\mathbf{1}
\left[
|\widehat q_i-q_i|>0.1R
\right].
\end{aligned}
\]

Metrics are averaged equally across coordinates, clips, and objects, in that order.

\end{document}